\documentclass[letterpaper, 10 pt, conference]{ieeeconf}
\IEEEoverridecommandlockouts
\usepackage{graphicx}
\usepackage{amsmath,amssymb,amsfonts}
\usepackage{booktabs}
\usepackage{float}
\usepackage{dblfloatfix}   % allow full-width figure*/table* at page bottom and fix their ordering, so
\usepackage{cite}
\floatstyle{ruled}
\newfloat{algorithm}{t}{loa}
\floatname{algorithm}{Algorithm}
\usepackage[colorlinks=true, linkcolor=blue, citecolor=blue, urlcolor=blue, breaklinks=true]{hyperref}
\hypersetup{pdftitle={SPACE: Swarm Pheromone Fields for Adaptive Collision-Aware Exploration},
  pdfauthor={Haohua Que, Haojia Gao, Mingkai Liu, Qian Zhang, Jiajun Sun, and Fei Qiao},
  pdfsubject={IEEE ROBIO 2026}, pdfkeywords={swarm robotics, exploration, stigmergy, collision avoidance}}

\title{\LARGE \bf
SPACE: Swarm Pheromone Fields for
Adaptive Collision-Aware Exploration}

\author{Haohua Que$^{1,\dagger}$, Haojia Gao$^{2,\dagger}$, Mingkai Liu$^{3}$, Qian Zhang$^{1}$, Jiajun Sun$^{1}$, and Fei Qiao$^{1,*}$%
\thanks{$^{\dagger}$These authors contributed equally. $^{*}$Corresponding author: Fei Qiao.}%
\thanks{$^{1}$Department of Electronic Engineering, Tsinghua University, Beijing, China. $^{2}$Tsinghua Shenzhen International Graduate School (SIGS), Tsinghua University, Shenzhen, China. $^{3}$School of Software and Microelectronics, Peking University, Beijing, China. {\tt\small qiaofei@tsinghua.edu.cn}}%
}

\begin{document}

\maketitle
\thispagestyle{empty}
\pagestyle{empty}

\begin{abstract}
Massive robot swarms can explore unknown environments quickly, but adding robots eventually stops helping. Doorways and dense traffic create congestion, increasing inter-robot contacts and reducing the value of each additional robot. We study this safety-efficiency tradeoff for ground swarms of tens to hundreds of robots. We present SPACE, Swarm Pheromone Fields for Adaptive Collision-Aware Exploration. Inspired by ant foraging, SPACE maintains a shared environmental field with an attractive frontier pheromone, a repellent explore pheromone, and a fast robot-density field. Robots use local field guidance on a synchronized shared map. We evaluate SPACE on real building floorplans, namely sixteen home layouts from the HouseExpo dataset and eight campus floors from the KTH dataset, with swarms of up to two hundred and fifty-six robots. At congested swarm sizes, SPACE attains the lowest inter-robot contact rate on HouseExpo and lies on the empirical Pareto frontier on KTH, together with handed lane formation at sizes where Lane has fewer contacts. Relative to a greedy nearest-frontier planner, SPACE reduces the contact rate by four to seventeen times while keeping coverage time within about two percent. The results indicate that, at this scale, coordination mainly improves safety rather than coverage time.
\end{abstract}

\section{Introduction}

Multi-robot exploration of an unknown environment is a core capability for search and rescue, inspection, and mapping \cite{cadena2016}. A common expectation is that a larger team explores faster. In practice, scale brings congestion. As the number of robots $N$ grows, robots crowd the same passages and frontiers, interfere with one another, and spend an increasing share of their time avoiding collisions rather than covering new ground. Studies of collective behavior describe a concave performance curve with an optimal swarm size, beyond which each additional robot adds congestion rather than coverage \cite{soma2023,fontllenas2018}.

\begin{figure}[t]
\centering
\includegraphics[width=\linewidth]{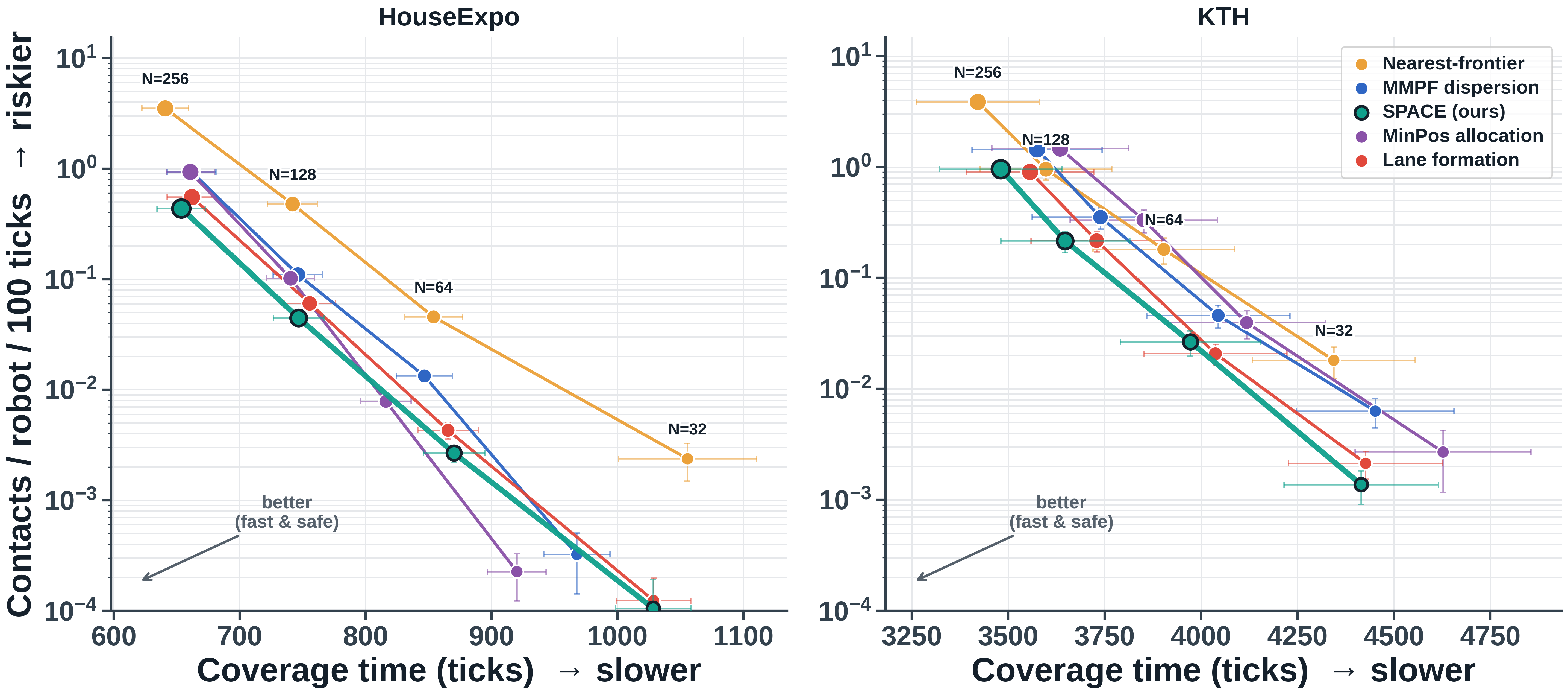}
\caption{Safety versus speed on HouseExpo (left) and KTH (right), with one standard error over map-and-seed runs. Lower-left is better. At congested sizes, SPACE has the lowest contact rate on HouseExpo. On KTH, SPACE and Lane trade safety for speed at $N{=}64,256$; SPACE dominates Lane at $N{=}128$.}
\label{fig:combined}
\vspace{-20pt}
\end{figure}

\begin{figure*}[t]
\centering
\includegraphics[width=\textwidth]{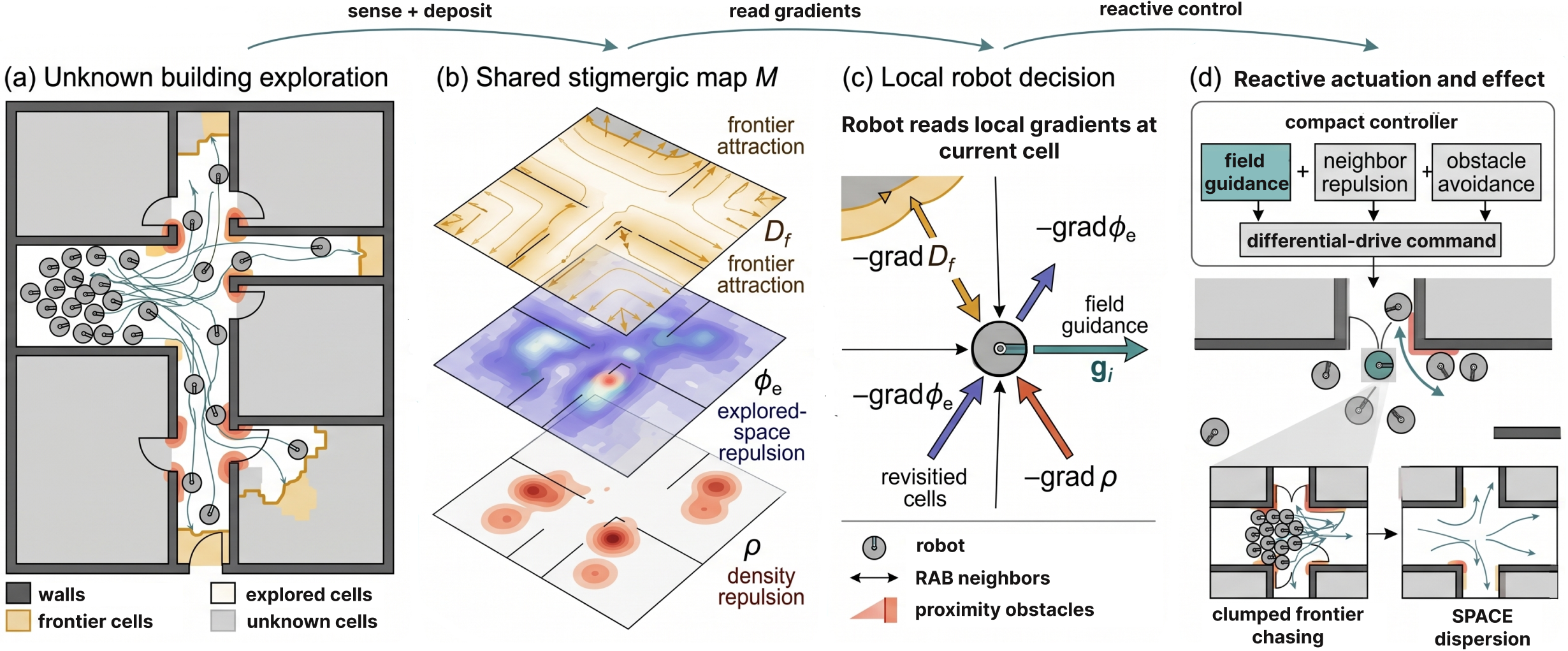}
\caption{Overview of SPACE. Robots explore an unknown building, write sensed cells and pheromone-like traces into a shared map, read local gradients of the frontier-distance, explore-pheromone, and density fields, and combine this field guidance with reactive neighbor and obstacle avoidance before moving. Local steering on this synchronized map preserves frontier-seeking behavior while dispersing robots away from revisited and crowded regions.}
\label{fig:overview}

\end{figure*}

At scale, exploration speed and safety trade off. Efficiency is the time to cover the reachable space. Safety is measured by an inter-robot proximity/contact rate; obstacle contacts are not evaluated. Figure~\ref{fig:combined} previews this tradeoff on the real-floorplan datasets we study. Three families of prior work optimize the first quantity and leave the second largely unmeasured. Frontier methods drive robots to the boundary between known and unknown space and report coverage time \cite{yamauchi,yamauchi98,burgard2005}. Sampling and next-best-view planners select informative viewpoints \cite{bircher2016}, and recent decentralized systems such as RACER and TARE coordinate a team or a single robot to high efficiency \cite{racer,tare}. Learning-based explorers train a policy that selects goals \cite{niroui2019}. All of these report efficiency, and standard exploration benchmarks standardize efficiency and cooperation metrics \cite{explorebench}, while inter-robot contact rate versus swarm size remains less commonly evaluated. This contact rate grows quickly as a swarm densifies, so it should be measured directly for massive swarms.

We take a biomimetic view of the coordination problem. Social insects coordinate large colonies without central control through stigmergy, the deposition and sensing of pheromones in the shared environment \cite{grasse1959,theraulaz1999,garnier2007}. Ants lay attractive trail pheromones toward food and repellent no-entry pheromones on unrewarding paths \cite{robinson2005}, and the signals evaporate so the colony stays current. For massive robot swarms, this environment-mediated pattern offers a compact alternative to explicit pairwise task allocation. SPACE adopts this mechanism. Its repellent field marks already-explored cells rather than unrewarding paths, so the analogy is functional rather than literal.

We present SPACE, Swarm Pheromone Fields for Adaptive Collision-Aware Exploration. SPACE maintains a shared field with an attractive frontier pheromone and a repellent pheromone, together with a short-lived robot-density field. Each robot reads only local gradients of this field and its own range-and-bearing sensor, and coordination is mediated only through the shared field. The repellent and density terms disperse robots away from explored and crowded regions, which reduces inter-robot contacts as the swarm scales. We evaluate SPACE in the ARGoS simulator \cite{argos} on real building floorplans, namely sixteen home layouts from HouseExpo \cite{houseexpo} and eight campus building floors from the KTH dataset \cite{aydemir2012}, with swarms of up to 256 ground robots. We compare against an uncoordinated nearest-frontier planner, a potential-field dispersion baseline in the spirit of MMPF \cite{smmr}, a MinPos-style frontier allocation \cite{minpos}, and a handed lane-formation rule. All methods share the same neighbor-repulsion and obstacle-avoidance parameters. Lane is the sole method that modifies the reactive layer, by adding a handed tangential term; the other comparisons isolate field guidance.
Our contributions are threefold.
\begin{enumerate}
\item \textbf{A safety and efficiency axis for massive-scale indoor exploration.} We push indoor swarm exploration on real floorplans to 256 ground robots, report the inter-robot contact rate explicitly against swarm size, and characterize the resulting safety-efficiency Pareto, which prior benchmarks omit.
\item \textbf{SPACE, a bio-inspired stigmergic coordinator.} A dual-pheromone field with local steering on a shared map, grounded in ant trail and no-entry pheromones, that keeps the contact rate low as the swarm scales at a small efficiency cost, and lies on the empirical Pareto frontier among the coordinated baselines we test.
\item \textbf{What coordination changes at scale.} We show that coordinated methods offer limited coverage-time gains over greedy nearest-frontier exploration in the congested regime; their main benefit is lower contact rate.
\end{enumerate}

% The paper is organized as follows. Section~\ref{sec:related} compares prior work. Section~\ref{sec:prelim} sets up the problem. Section~\ref{sec:method} presents SPACE. Section~\ref{sec:setup} gives the experimental setup, Section~\ref{sec:results} the results, and Section~\ref{sec:disc} discusses limitations.

\section{Related Work}
\label{sec:related}

\textbf{Multi-robot exploration.} Frontier-based exploration is efficient and simple, but a nearest-frontier rule lets robots converge on the same boundary and interfere \cite{yamauchi,yamauchi98}. Coordinated exploration assigns robots to frontiers to reduce overlap \cite{burgard2005,minpos,simmons2000,zlot2002}, information-gain and next-best-view planners select informative viewpoints \cite{stachniss2005,bircher2016}, and decentralized active-perception planners coordinate teams online \cite{best2019}. RACER coordinates a decentralized team through online space decomposition at high efficiency, but it is aerial and validated at up to ten robots, and it does not report how safety scales with team size \cite{racer}. TARE achieves high efficiency but is single-robot \cite{tare}. Learning-based explorers select goals with a trained policy \cite{niroui2019}, and benchmarks standardize the efficiency metrics \cite{explorebench}. Most of this work is validated in synthetic or small layouts. We instead evaluate on real building floorplans from HouseExpo \cite{houseexpo} and the KTH dataset \cite{aydemir2012}, at hundreds of ground robots, and we measure the safety axis directly.

\textbf{Swarm and stigmergic coordination.} Swarm robotics seeks scalable collective behavior from simple local rules \cite{sahin2005,brambilla2013,hamann2018}. Stigmergy, coordination through traces left in a shared environment, underlies swarm intelligence broadly \cite{bonabeau1999,garnier2007}, ant colony optimization \cite{dorigo2006}, and pheromone robotics \cite{payton2001,fontllenas2018}. Potential-field and dispersion methods coordinate coverage through a shared signal. MMPF disperses a team with a per-robot repulsive potential so robots explore distinct areas \cite{smmr}, which improves allocation but does not by itself bound the collision rate at scale. A real-robot study shows that strong repellent stigmergy can reverse and perform worse than a random walk at high density \cite{hunt2019}. SPACE differs by combining an attractive frontier pheromone, a stigmergic explore pheromone that suppresses redundant revisiting, and a fast density field, and by pairing the field with a light reactive layer so that close-range safety does not depend on the repellent field alone.

\textbf{Collision avoidance and congestion.} Reactive multi-robot collision avoidance is well studied. Velocity-obstacle and reciprocal velocity-obstacle methods produce smooth avoidance for many agents \cite{fiorini1998,rvo,orca}, control barrier functions add safety guarantees for multi-robot systems \cite{ames2019,wang2017}, and buffered Voronoi cells give decentralized, communication-light safety \cite{buavc}. Discrete multi-agent path finding solves the same problem on a graph with known goals \cite{sharon2015,okumura2019}. Congestion-aware planners embed flow penalties for large-scale navigation to a known goal \cite{cmpp,cglr}, and coverage control deploys a team over a known region with potential fields \cite{cortes2004,howard2002,schwager2009}. These methods assume a known goal or region and address navigation rather than exploration of unknown space. SPACE borrows their reactive spirit for the close-range layer, with common repulsion parameters and an additional tangential term for Lane, and uses the shared stigmergic field to disperse the team.

\section{Preliminaries and Problem Setup}
\label{sec:prelim}

\textbf{Environment and task.} A team of $N$ homogeneous ground robots explores an initially unknown bounded planar environment $\mathcal{W}\subset\mathbb{R}^2$, partitioned into free space $\mathcal{W}_{\mathrm{free}}$ and obstacles. The environment is a building floorplan, a set of rooms joined by doorways. We discretize it into an occupancy grid of square cells, each labeled \textsc{unknown}, \textsc{free}, or \textsc{obstacle}. The task is to cover the reachable free space, that is to raise the known free area to a target fraction (95 percent) of the free area reachable from the start, beginning from an initial cluster of robots.

\textbf{Robot and sensing.} Each robot is a differential-drive disk of radius $r$ with state $(\mathbf{p}_i,\theta_i)$, position and heading, and a bounded wheel speed. It carries three sensors. A positioning sensor gives its own pose. A range-and-bearing (RAB) sensor reports the range and bearing of neighbors within a finite range. A ring of proximity sensors reports nearby obstacles. The robot perceives occupancy within a sensing radius $r_s$, which reveals ground-truth cell labels within that radius without ray casting or occlusion. Proximity sensing separately handles local obstacle avoidance.

\textbf{Shared map as a stigmergic medium.} We assume the team maintains a shared occupancy grid $M$, as provided by a collaborative mapping backend, and we treat $M$ as the shared environment on which the swarm leaves traces. SPACE is a coordination layer that writes and reads scalar fields on $M$. A robot samples field gradients around its own cell. In our implementation, centralized simulator loop functions maintain the synchronized map, update the fields, and periodically run global BFS. Steering is local, but the mapping backend is not a fully decentralized implementation; communication delay and bandwidth are not modeled.

\textbf{Objective and metrics.} We measure efficiency by the coverage time $T_{\mathrm{cov}}$, the number of control ticks to reach the coverage target. We measure safety by the inter-robot contact rate, the time-averaged number of robot pairs within a contact clearance, normalized per robot. The goal is to improve the empirical Pareto frontier in the $(T_{\mathrm{cov}},\,\text{contact rate})$ plane as $N$ grows. Section~\ref{sec:setup} gives the operational definitions.

\section{Method}
\label{sec:method}

SPACE has two layers. A shared stigmergic field carries global coordination across the team, and a local reactive controller provides close-range safety. Figure~\ref{fig:overview} summarizes the closed-loop system. The two layers are deliberately separated. The field decides where to go and disperses the team, while the reactive layer keeps robots apart at contact range, so the method never relies on the field alone when the swarm is dense. Algorithm~\ref{alg:space} states one simulation tick. We describe each layer in turn.

\subsection{Layer 1: shared map and three fields}
The team maintains a shared occupancy grid $M$ over the arena, with each cell marked unknown, free, or obstacle. A robot reveals the cells within its sensing radius and writes them into $M$. From $M$ we maintain three scalar fields on the grid, one for efficiency and two for safety.

\emph{Frontier-distance field $D_f$ (efficiency).} Frontier cells are free cells adjacent to unknown cells. $D_f$ is the breadth-first distance from the nearest frontier over all cells except known obstacles, recomputed every $T_f$ ticks by a multi-source search from all frontier cells. Unknown cells are optimistically traversable, so descending $-\nabla D_f$ supplies frontier guidance rather than a guaranteed feasible path through known free space.

\emph{Explore pheromone $\phi_e$ (anti-redundancy).} Each sensed free cell receives a unit deposit, and the field decays by a factor $\gamma_e$ per tick,
\begin{equation}
\phi_e \leftarrow \gamma_e\,\phi_e, \qquad \phi_e(c) \mathrel{+}= 1 \;\; \forall c \in \mathcal{S}_i,
\end{equation}
where $\mathcal{S}_i$ contains the free cells sensed by robot $i$. Decay precedes all robots' deposits each tick. High $\phi_e$ marks explored space, and $-\nabla \phi_e$ repels robots from it. This is the no-entry analogue, and it suppresses redundant revisiting.

\emph{Density field $\rho$ (crowding).} Each robot deposits at its own cell, and $\rho$ decays quickly with $\gamma_\rho \ll \gamma_e$,
\begin{equation}
\rho \leftarrow \gamma_\rho\,\rho, \qquad \rho(c_i) \mathrel{+}= 1,
\end{equation}
so $\rho$ tracks momentary crowding. The term $-\nabla \rho$ disperses robots away from dense regions before contacts form.

\subsection{Layer 1 output: decentralized field guidance}
Robot $i$ at cell $c_i$ reads only the local gradients of the three fields at its own cell and combines them into a world-frame steering vector,
\begin{equation}
\mathbf{g}_i = w_f\,\widehat{(-\nabla D_f)} + w_e\,\widehat{(-\nabla \phi_e)} + w_\rho\,\widehat{(-\nabla \rho)},
\label{eq:guidance}
\end{equation}
where $\widehat{\mathbf{v}}=\mathbf{v}/\|\mathbf{v}\|$ for $\|\mathbf{v}\|>10^{-6}$ and zero otherwise. The fixed weights $(w_f,w_e,w_\rho)=(1.0,0.4,0.9)$ retain frontier attraction while adding repulsion. We use one setting across layouts and swarm sizes to compare coordination without layout-specific tuning; it is not claimed optimal for every building. Each gradient uses four-neighbor finite differences. Out-of-grid or infinite-distance samples contribute as flat neighbors. Reactive sensing handles obstacles; the field alone does not guarantee collision avoidance.

\subsection{Layer 2: reactive safety and actuation}
On top of the field guidance, each robot adds a short-range reactive term from its range-and-bearing sensor and its proximity sensor,
\begin{equation}
\mathbf{u}_i = R(-\theta_i)\,\widetilde{\mathbf{g}}_i + w_n \sum_{j \in \mathcal{N}_i} a(d_{ij})\,(-\hat{\mathbf{b}}_{ij}) + w_o\,\mathbf{r}_i,
\label{eq:reactive}
\end{equation}
where $R(\alpha)$ is the standard counterclockwise planar rotation, so $R(-\theta_i)$ maps world coordinates into the robot frame. The controller uses $\widetilde{\mathbf{g}}_i=\mathbf{g}_i/\|\mathbf{g}_i\|$ for $\|\mathbf{g}_i\|>10^{-3}$ and zero otherwise. Here $\mathcal{N}_i$ are neighbors within range, $\hat{\mathbf{b}}_{ij}$ is the bearing to neighbor $j$, $a(d_{ij})=\max(0,1-d_{ij}/d_n)$ is a range-decreasing weight that vanishes beyond a personal-space radius $d_n$, and $\mathbf{r}_i$ is an obstacle-avoidance vector from proximity. The neighbor term is the close-range collision-avoidance layer, and its strength $w_n$ is the single shared safety parameter. Let $e_i$ be the angle of $\mathbf{u}_i$ in the robot frame. The forward and turning wheel-speed components are $v\,\max(0,\cos e_i)$ and $v\,\tanh(2 e_i)$, and the left and right wheel speeds are their difference and sum, clamped to $[-v,v]$. A robot facing away from its target turns in place before advancing, rather than forcing through an obstacle. If $\|\mathbf{u}_i\|<10^{-4}$, both wheels move forward at $0.3v$. Local actuation uses the shared field and onboard sensing, without explicit pairwise task allocation.

All methods share $w_n$, $d_n$, $w_o$, and the wheel controller. Frontier sets $w_e{=}w_\rho{=}0$, the MMPF-style density-only adaptation sets $w_e{=}0$, and SPACE uses all three terms. The MinPos-style adaptation replaces frontier guidance with per-region allocation and retains density repulsion. Lane uses SPACE's field and is the sole exception to identical Layer 2: it adds a right-handed tangential neighbor term of weight $0.8$. Thus its comparison evaluates this additional traffic rule.

\begin{algorithm}[t]
\caption{SPACE: one simulation tick}
\label{alg:space}
\small
\noindent\textbf{Shared:} occupancy grid $M$; fields $D_f, \phi_e, \rho$; tick $t$\par
\smallskip
\textit{// Layer 1a: decay, then sense and deposit}\par
$\phi_e \leftarrow \gamma_e\,\phi_e$; \quad $\rho \leftarrow \gamma_\rho\,\rho$\par
\textbf{for} each robot $i$ \textbf{do}\par
\hspace*{1.3em} reveal cells within sensing radius; update $M$\par
\hspace*{1.3em} $\mathcal{S}_i \leftarrow$ sensed free cells of robot $i$\par
\hspace*{1.3em} $\phi_e(c) \mathrel{+}= 1 \;\, \forall c \in \mathcal{S}_i$; \quad $\rho(c_i) \mathrel{+}= 1$\par
\textbf{if} $t \bmod T_f = 0$ \textbf{then} recompute $D_f$ by multi-source BFS\par
\smallskip
\textit{// Layer 1b + Layer 2: act, decentralized and in parallel}\par
\textbf{for} each robot $i$ \textbf{in parallel do}\par
\hspace*{1.3em} $\mathbf{g}_i \leftarrow w_f\widehat{(-\nabla D_f)} + w_e\widehat{(-\nabla \phi_e)} + w_\rho\widehat{(-\nabla \rho)}$\par
\hspace*{1.3em} $\mathbf{u}_i \leftarrow R(-\theta_i)\widetilde{\mathbf{g}}_i + w_n\!\sum_{j\in\mathcal{N}_i} a(d_{ij})(-\hat{\mathbf{b}}_{ij}) + w_o\mathbf{r}_i$\par
\hspace*{1.3em} actuate wheels from $\mathbf{u}_i$ using Layer 2 rules\par
$t \leftarrow t + 1$
\end{algorithm}

\section{Experimental Setup}
\label{sec:setup}

\textbf{Simulator and robot.} We use ARGoS \cite{argos} with the two-dimensional dynamics engine and the foot-bot model, with differential steering, a range-and-bearing sensor, proximity sensors, and positioning. The shared fields, metrics, and baselines are implemented in C++, with Python experiment and plotting scripts. The code is open-source at \href{https://github.com/SenseLabRobo4111/SPACE}{github.com/SenseLabRobo4111/SPACE}.

\begin{figure*}[tb]
\centering
\includegraphics[width=\textwidth]{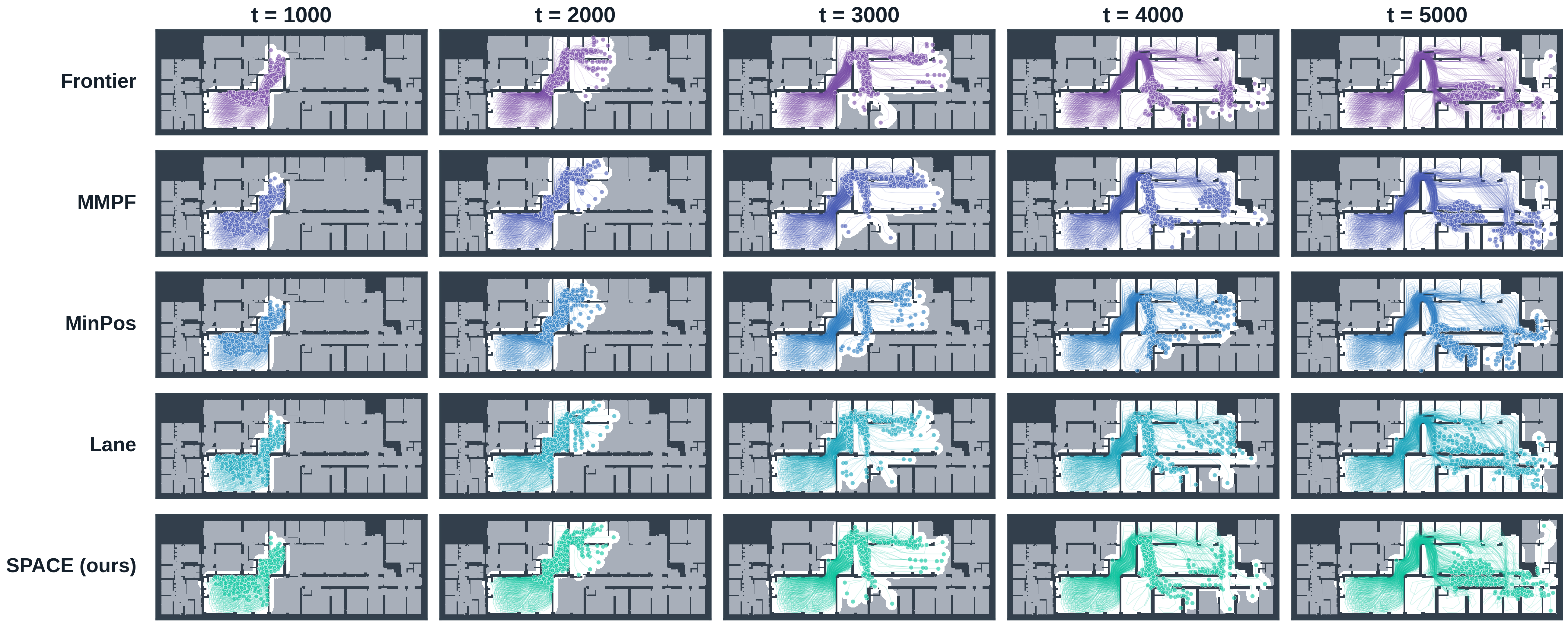}
\caption{Robot trajectories over the course of exploration on a large KTH campus floor (the Valhallav\"agen floor \texttt{50010539}, a $63\times23$\,m, $52$-room building, $N{=}256$). Rows are the five methods, columns are five instants (control ticks $1000$ to $5000$, left to right). Each thin line is one robot's path so far, and transparency makes high-traffic regions darker. White is explored, gray is not yet explored, walls are dark. The greedy nearest-frontier planner (top) keeps the swarm bunched in a dense clump that crawls outward, while SPACE (bottom) spreads along the corridors and into the rooms. The resulting dispersion lowers the contact rate as the swarm scales (Table~\ref{tab:main}).}
\label{fig:explore}
\end{figure*}

\begin{figure}[tb]
\centering
\includegraphics[width=\linewidth]{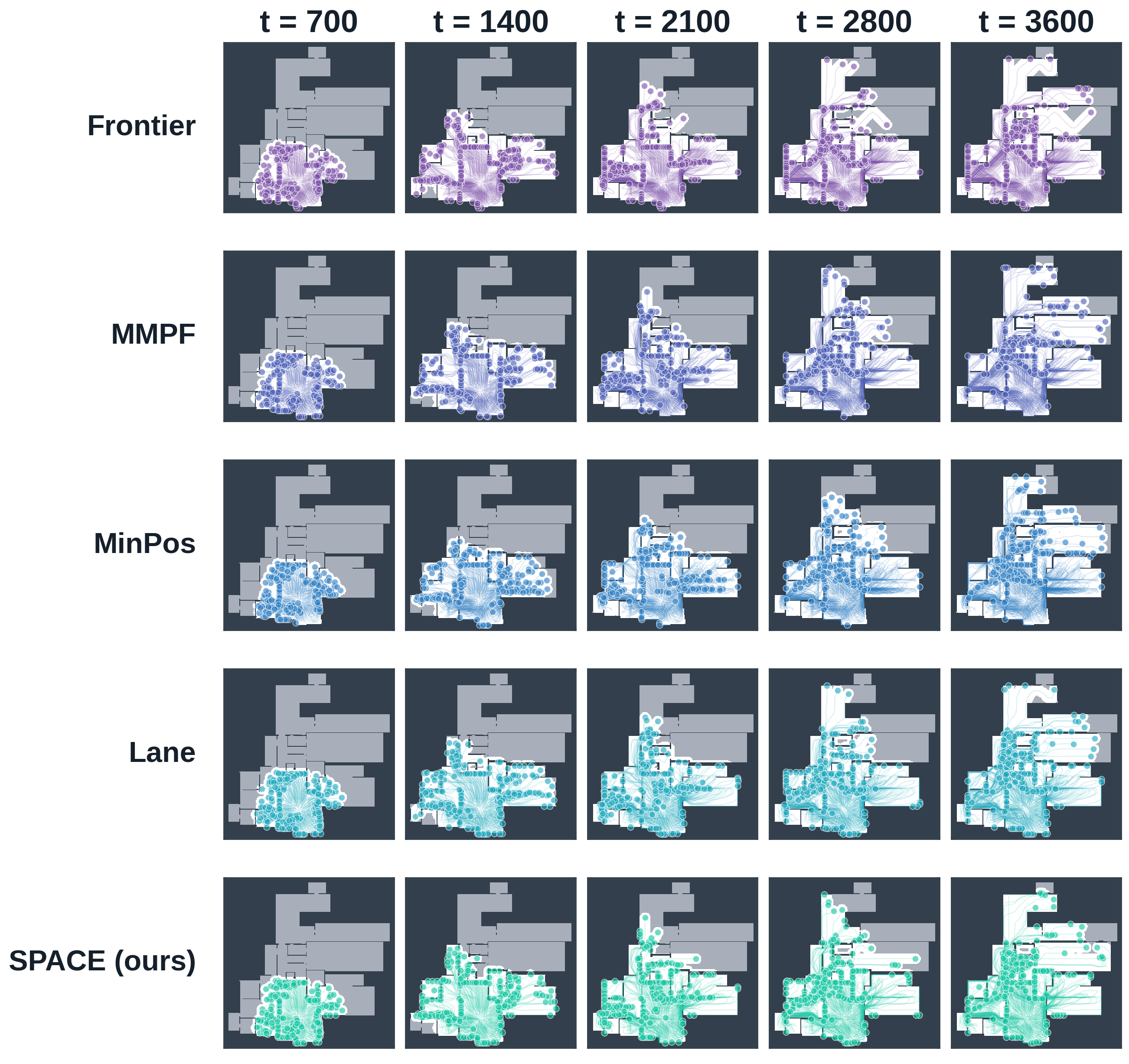}
\caption{Robot trajectories on a large HouseExpo home (map \texttt{847e043c}, a $45\times45$\,m, $30$-room layout, $N{=}256$), in the same format as Fig.~\ref{fig:explore}. Rows are the five methods, columns are five instants (control ticks $700$ to $3600$, left to right). The greedy nearest-frontier planner (top) keeps the swarm packed near the deployment room and works through the home slowly, while SPACE (bottom) disperses across the rooms early. The same clump-versus-disperse contrast holds in the cluttered homes as on the open KTH corridors.}
\label{fig:explore_he}
\end{figure}

\textbf{Environments.} We evaluate on real building floorplans. From HouseExpo \cite{houseexpo} we take sixteen home layouts, and from the KTH dataset \cite{aydemir2012} we take eight campus building floors of 25 to 70 meters across. Each floorplan is rasterized into an occupancy grid, with doorways carved open so that rooms connect into one building, and the swarm starts clustered in the most open free region and must explore outward. The doorways are the congestion bottleneck, and the two datasets span a range from small cluttered homes to long-corridor campus floors. Figures~\ref{fig:explore} and~\ref{fig:explore_he} show how the methods explore these two environment types, and we return to them in the results.

\textbf{Swarm sizes and seeds.} We sweep $N$ from 16 to 256 robots, with 20 random seeds per configuration on the sixteen HouseExpo homes and 8 on the eight KTH floors. We report means and one standard error over map-and-seed runs (320 HouseExpo or 64 KTH runs per method and size), consistent with Fig.~\ref{fig:combined}. Common parameters appear in Table~\ref{tab:params}.

\textbf{Metrics.} Coverage time is the first tick reaching 95 percent of reachable free cells; time-outs enter the mean at the run cap (16,000 ticks for HouseExpo, 40,000 for KTH). After excluding the first 50 ticks, contact rate is $100\sum_t C_t/[N(T-50)]$, where $C_t$ counts unordered robot pairs with center distance below $0.22$\,m and $T$ is run length. A pair remaining close contributes once each tick. This measures proximity duration, not distinct physical collisions; obstacle contacts are not counted. Travel is total team path length divided by the number of covered free cells (m/cell). At $N\geq64$, all runs reach the target. On HouseExpo, Frontier times out in 28/320 runs at $N{=}16$ and 1/320 at $N{=}32$; Lane times out in 1/320 at $N{=}16$. All other runs succeed.

\textbf{Baselines.} \emph{Frontier} is an uncoordinated nearest-frontier planner. \emph{MMPF} is a density-only adaptation inspired by \cite{smmr}, omitting the explore pheromone. \emph{MinPos} (code: Alloc) adapts frontier allocation \cite{minpos}. \emph{Lane} (code: Flow) adds handed avoidance. These are controlled adaptations, not full reproductions of the referenced systems. SPACE is the full field method.

\begin{table}[t]
\centering
\caption{Simulation parameters and metric conventions. Parameters are shared across methods except the field terms and Lane's tangential term.}
\label{tab:params}
\small
\setlength{\tabcolsep}{3pt}
\renewcommand{\arraystretch}{1.08}
\begin{tabular}{@{}ll@{}}
\toprule
Parameter & Value \\
\midrule
Nominal grid cell side & $0.20$\,m (HE); $0.15$\,m (KTH) \\
Control period & $0.1$\,s (10 ticks/s) \\
Foot-bot body radius; wheel limit & $0.085$\,m; $v=0.15$\,m/s \\
Map sensing; RAB range & $1.5$\,m; $3.0$\,m \\
Neighbor avoidance $d_n$; $w_n,w_o$ & $0.6$\,m; $2.0,1.5$ \\
SPACE weights $w_f,w_e,w_\rho$ & $1.0,0.4,0.9$ \\
Per-tick decay $\gamma_e,\gamma_\rho$ & $0.999,0.55$ \\
Frontier refresh $T_f$ & 4 ticks ($0.4$\,s) \\
Contact center-distance threshold & $0.22$\,m \\
Excluded startup; coverage target & 50 ticks ($5$\,s); $95\%$ \\
\bottomrule
\end{tabular}
\end{table}

\section{Results}
\label{sec:results}

\subsection{Safety and efficiency across swarm size}
Figure~\ref{fig:combined} places each method in the coverage-time versus contact-rate plane as the swarm grows, on both datasets. At congested sizes ($N{=}64,128,256$), SPACE has the lowest contact rate on HouseExpo. On KTH, Lane has a lower contact rate at $N{=}64,256$, while SPACE covers faster; SPACE dominates Lane at $N{=}128$. The contact rate of every method rises with $N$ as robots crowd doorways, but SPACE rises far more slowly, so its absolute contact-rate advantage over the greedy nearest-frontier planner widens with swarm size. On the HouseExpo homes the frontier planner reaches a contact rate of $0.046$, $0.48$, and $3.50$ at $N{=}64$, $128$, and $256$, while SPACE stays at $0.003$, $0.044$, and $0.435$, which is between eight and seventeen times fewer contacts. On the KTH building floors the same comparison is $0.18$, $0.95$, and $3.86$ for the frontier planner against $0.026$, $0.215$, and $0.954$ for SPACE, four to seven times fewer. Table~\ref{tab:main} collects these gains. The ratio narrows as $N$ grows, but the absolute benefit increases: on the HouseExpo homes the close encounters avoided by SPACE rise from $0.04$ to $3.07$ per robot per 100 ticks between $N{=}64$ and $N{=}256$. This safety margin costs little coverage time. Measured against the greedy nearest-frontier baseline, SPACE is faster when the swarm is small, where greedy chasing of the nearest boundary makes robots redundant, for example $1565$ ticks versus $2523$ at $N{=}16$ on HouseExpo. It stays within about two percent at the congested sizes, for example $654$ ticks versus $641$ at $N{=}256$. On KTH, SPACE is within two percent of the frontier planner at congested sizes while remaining far safer. No tested method is both faster and safer than SPACE at any congested size. Table~\ref{tab:main} reports each method's contact rate alongside SPACE's coverage time and safety gain. The gains are consistent across runs: SPACE has no more contacts than the greedy planner on 90 to 100 percent of individual map-and-seed pairs, and each congested-size SPACE versus Frontier contact-rate comparison gives $p<0.001$ in a two-sided paired $t$-test. Pairs share a map and seed, pooling all runs within each dataset and size. These run-level tests reuse maps and do not establish significance over independent new buildings.

\begin{table}[t]
\centering
\caption{Main results on real building floorplans, mean over maps and seeds. Left: inter-robot contact rate (robot pairs within a contact clearance, per robot per 100 ticks; lower is safer, minimum before rounding in bold). Right, for SPACE: coverage time to 95 percent coverage, its percentage difference from the greedy nearest-frontier baseline ($\Delta$, negative is faster), and the safety gain, the factor by which SPACE lowers the greedy contact rate. No method is both faster and safer than SPACE at any congested size.}
\label{tab:main}
\setlength{\tabcolsep}{5pt}
\renewcommand{\arraystretch}{1.2}
\resizebox{\columnwidth}{!}{%
\begin{tabular}{@{}lrrrrrrrc@{}}
\toprule
& \multicolumn{5}{c}{Contact rate (lower safer)} & \multicolumn{3}{c}{SPACE eff.\ \& safety} \\
\cmidrule(lr){2-6}\cmidrule(lr){7-9}
$N$ & Front. & MMPF & MinP & Lane & \textbf{SPACE} & Time & $\Delta$ & gain \\
\midrule
\multicolumn{9}{@{}l}{\textit{HouseExpo: 16 homes, 20 seeds}}\\
16  & 0.000 & \textbf{0.000} & \textbf{0.000} & \textbf{0.000} & \textbf{0.000} & 1565 & $-38\%$ & -- \\
32  & 0.002 & 0.000 & 0.000 & 0.000 & \textbf{0.000} & 1028 & $-3\%$  & -- \\
64  & 0.046 & 0.013 & 0.008 & 0.004 & \textbf{0.003} & 870  & $+2\%$  & \textbf{17.1}$\times$ \\
128 & 0.477 & 0.110 & 0.102 & 0.060 & \textbf{0.044} & 747  & $+1\%$  & \textbf{10.7}$\times$ \\
256 & 3.501 & 0.932 & 0.932 & 0.551 & \textbf{0.435} & 654  & $+2\%$  & \textbf{8.1}$\times$ \\
\midrule
\multicolumn{9}{@{}l}{\textit{KTH: 8 campus floors, 8 seeds}}\\
16  & 0.001 & 0.000 & 0.001 & \textbf{0.000} & 0.000 & 5113 & $-3\%$ & -- \\
32  & 0.018 & 0.006 & 0.003 & 0.002 & \textbf{0.001} & 4415 & $+2\%$ & -- \\
64  & 0.181 & 0.046 & 0.040 & \textbf{0.021} & 0.026 & 3972 & $+2\%$ & \textbf{6.8}$\times$ \\
128 & 0.953 & 0.353 & 0.332 & 0.217 & \textbf{0.215} & 3648 & $+1\%$ & \textbf{4.4}$\times$ \\
256 & 3.855 & 1.439 & 1.464 & \textbf{0.901} & 0.954 & 3481 & $+2\%$ & \textbf{4.0}$\times$ \\
\bottomrule
\end{tabular}}
\end{table}

\subsection{Why the field, and which baseline it beats}
The stigmergic explore pheromone separates SPACE from the MMPF dispersion baseline. MMPF disperses with a density term alone, whereas SPACE adds the explore pheromone that suppresses redundant revisiting. The MMPF versus SPACE columns of Table~\ref{tab:main} isolate this one term. Adding the explore pheromone to the density-only baseline lowers the contact rate by a factor of $2.1$ to $5.0$ on the HouseExpo homes and $1.5$ to $1.8$ on the KTH floors, at every congested swarm size, because it spreads the team over distinct regions and reduces the close encounters that revisiting creates. Figures~\ref{fig:explore} and~\ref{fig:explore_he} visualize the mechanism on a KTH floor and a HouseExpo home. The uncoordinated planner keeps the swarm in a tight clump that crawls outward, while SPACE disperses the swarm along the corridors and into the rooms.

\subsection{What coordination changes at scale}
The greedy nearest-frontier rule has competitive coverage time in the congested regime. At $N{=}256$, it is the fastest evaluated method on both datasets; allocation provides no consistent speed advantage across congested sizes. The coordinated methods mainly improve dispersion and contact rate. Handed lane formation helps only at the margin. On the long corridors of the KTH floors, lane formation reaches a contact rate close to SPACE, for example $0.90$ against $0.95$ at $N{=}256$, but at a higher coverage time. SPACE and Lane are jointly Pareto-optimal on KTH at $N{=}64,256$; SPACE dominates Lane at $N{=}128$ and on HouseExpo at $N{=}128,256$. At HouseExpo $N{=}64$, Lane is slightly faster but has more contacts. Across congested sizes, SPACE's mean path length per covered cell differs from Frontier by $-3.3\%$ to $+7.4\%$ on HouseExpo and $+1.6\%$ to $+2.1\%$ on KTH. These modest travel changes accompany much larger contact-rate reductions. Acceleration, turning time, and energy use were not measured separately, so path length does not rule out additional turning or idling costs. Across both datasets, coordination mainly reduces contact rather than coverage time.

\section{Discussion and Limitations}
\label{sec:disc}

The main result is that, in these dense indoor swarms, coordination improves safety more than it improves coverage time. At congested sizes, SPACE has the lowest contact rate on HouseExpo and remains on the empirical Pareto frontier on KTH, where Lane offers a safety-speed tradeoff at $N{=}64,256$. SPACE's coverage time stays within about two percent of the greedy frontier baseline; no optimal-time lower bound is claimed.

The safety advantage is most useful for large teams in confined structures, where doorways and corridors concentrate traffic. In sparsely populated buildings, contact rates and the margins among coordinated methods diminish.

Field sampling uses constant-size neighborhoods, but shared field decay and BFS scale with grid size, and reactive processing scales with sensed neighbors. Runtime and communication scaling of a distributed backend remain unevaluated.

This study is in simulation. Radius-based ground-truth map reveal ignores occlusion, and optimistic BFS can route guidance through unknown space. These shared assumptions may understate sensing difficulty and alter frontier access; visibility-aware sensing and hardware trials are needed. The 95 percent coverage target can leave difficult corners or narrow spaces unexplored, so it does not establish complete exploration. Fixed weights may require adjustment in other geometries, and an explore-only ablation remains future work. Future work can seed the frontier field with a learned predictor of unseen structure, so the team is pulled toward likely rooms before it senses them, and extend the planar field to three dimensions for aerial swarms.

\section{Conclusion}

We presented SPACE, a stigmergic dual-pheromone field for massive-scale swarm exploration, grounded in ant trail and no-entry pheromones. At congested swarm sizes, SPACE has the lowest inter-robot contact rate on the sixteen HouseExpo homes and lies on the empirical Pareto frontier on the eight KTH floors, jointly with Lane at $N{=}64,256$. Relative to greedy frontier exploration, it reduces contact rate by four to seventeen times at a coverage-time cost of about two percent or less. These results support field-mediated coordination for reducing inter-robot proximity in large indoor swarms.

\section*{Acknowledgment}
The authors would like to acknowledge support from the Beijing Natural Science Foundation (L253009).

\end{document}